\newcommand{\parskiny}{}
\newcommand{\seckiny}{}
\newcommand{\subseckiny}{}
\newcommand{\figskiny}{}
\newcommand{\eqskiny}{}

\newcommand{\done}[1]{}
\newcommand{\matr}[1]{\mathbf{#1}} 

\documentclass{article}

\PassOptionsToPackage{numbers,sort}{natbib}
\usepackage[final]{nips_2016}

\usepackage[utf8]{inputenc} 
\usepackage[T1]{fontenc}    
\usepackage{hyperref}       
\usepackage{url}            
\usepackage{booktabs}       
\usepackage{amsfonts}       
\usepackage{nicefrac}       
\usepackage{microtype}      
\usepackage{graphicx}
\usepackage{color}
\usepackage{subfigure}
\usepackage{algorithm}
\usepackage{algpseudocode}

\title{Strategic Attentive Writer for Learning Macro-Actions}

\author{
  Alexander (Sasha) Vezhnevets, Volodymyr Mnih, John Agapiou, \\ 
  \textbf{Simon Osindero, Alex Graves, Oriol Vinyals, Koray Kavukcuoglu}\\
  Google DeepMind\\
  \texttt{\{vezhnick,vmnih,jagapiou,osindero,gravesa,vinyals,korayk\}@google.com} \\
}

\begin{document}

\maketitle

\begin{abstract}
\vspace{-2mm}

We present a novel deep recurrent neural network architecture that learns to build \emph{implicit plans} in an end-to-end manner by purely interacting with an environment in reinforcement learning setting. The network builds an internal plan, which is continuously updated upon observation of the next input from the environment. It can also partition this internal representation into contiguous sub- sequences by learning for how long the plan can be committed to -- i.e. followed without re-planing. Combining these properties, the proposed model, dubbed STRategic Attentive Writer (STRAW) can learn high-level, temporally abstracted macro- actions of varying lengths that are solely learnt from data without any prior information.  These macro-actions enable both structured exploration and economic computation. We experimentally demonstrate that STRAW delivers strong improvements on  several ATARI games by employing temporally extended planning strategies (e.g. Ms. Pacman and Frostbite). It is at the same time a general algorithm that can be applied on any sequence data. To that end, we also show that when trained on text prediction task, STRAW naturally predicts frequent n-grams (instead of macro- actions), demonstrating the generality of the approach.

\end{abstract}

\seckiny
\section{Introduction}
\seckiny

Using reinforcement learning to train neural network controllers has recently led to rapid progress on a number of challenging control tasks~\cite{mnih-dqn-2015,schulman2015trust,levine2015endtoend}.
Much of the success of these methods has been attributed to the ability of neural networks to learn useful abstractions or representations of the stream of observations, allowing the agents to generalize between similar states.
Notably, these agents do not exploit another type of structure -- the one present in the space of controls or policies.
Indeed, not all sequences of low-level controls lead to interesting high-level behaviour and an agent that can automatically discover useful macro-actions should be capable of more efficient exploration and learning.
The discovery of such temporal abstractions has been a long-standing problem in both reinforcement learning and sequence prediction in general, yet no truly scalable and successful architectures exist.

We propose a new deep recurrent neural network architecture, dubbed STRategic Attentive Writer (STRAW), that is capable of learning macro-actions in a reinforcement learning setting.
Unlike the vast majority of reinforcement learning approaches~\cite{mnih-dqn-2015,schulman2015trust,levine2015endtoend}, which output a single action after each observation, STRAW maintains a multi-step action plan.
STRAW periodically updates the plan based on observations and commits to the plan between the replanning decision points.
The replanning decisions as well as the commonly occurring sequences of actions, i.e. macro-actions, are learned from rewards.
To encourage exploration with macro-actions we introduce a noisy communication channel between a feature extractor (e.g. convolutional neural network) and the planning modules, taking inspiration from recent developments in variational auto-encoders~\cite{kingma2014auto,rezende2014stochastic,gregor2015draw}. Injecting noise at this level of the network generates randomness in plans updates that cover multiple time steps and thereby creates the desired effect.

Our proposed architecture is a step towards more natural decision making, wherein one observation can generate a whole sequence of outputs if it is informative enough.
This provides several important benefits.
First and foremost, it facilitates structured exploration in reinforcement learning -- as the network learns meaningful action patterns it can use them to make longer exploratory steps in the state space~\cite{botvinick2009hierarchically}.
Second, since the model does not need to process observations while it is committed its action plan, it learns to allocate computation to key moments thereby freeing up resources when the plan is being followed.
Additionally, the acquisition of macro-actions can aid transfer and generalization to other related problems in the same domain (assuming that other problems from the domain share similar structure in terms of action-effects).

We evaluate STRAW on a subset of Atari games that require longer term planning and show that it leads to substantial improvements in scores.
We also demonstrate the generality of the STRAW architecture by training it on a text prediction task and show that it learns to use frequent n-grams as the macro-actions on this task.

The following section reviews the related work. Section~\ref{sec:model} defines the STRAW model formally. Section~\ref{sec:learning} describes the training procedure for both supervised and reinforcement learning cases. Section~\ref{sec:experiments} presents the experimental evaluation of STRAW on 8 ATARI games, 2D maze navigation and next character prediction tasks. Section~\ref{sec:conclusion} concludes.

\section{Related Work}
\label{sec:related_work}
\seckiny
Learning temporally extended actions and temporal abstraction in genral are long standing problems in reinforcement learning~\cite{sutton1999between,precup2000temporal, dayan1993feudal, dietterich2000hierarchical, boutilier1997prioritized, dayan1993improving, kaelbling2014hierarchical, parr1998reinforcement, precup1997planning, precup1998theoretical, schmidhuber1991neural, sutton1995td}. The options framework~\cite{sutton1999between,precup2000temporal} provides a general formulation. An option is a sub-policy with a termination condition, which takes in enviroment observations and outputs actions until the termination condition is met. 
An agent picks an option using its policy-over-options and subsequently follows it until termination, at which point the policy-over-options is queried again and the process continues. 
Notice, that macro-action is a particular, simpler instance of options, where the action sequence (or a distribution over them) is decided at the time the macro-action is initiated.
Options are typically learned using subgoals and 'pseudo-rewards' that are provided explicitly~\cite{sutton1999between, dietterich2000hierarchical, dayan1993feudal}. Given the options, a policy-over-options can be learned using standard techniques by treating options as actions. Recently~\cite{tessler2016minecraft, tejas2016hdrl} have demonstrated that combining deep learning with pre-defined subgoals delivers promising results in challenging environments like Minecraft and Atari, however, subgoal discovery remains an unsolved problem.
Another recent work by~\cite{bacon2015option} shows a theoretical possibility of learning options jointly with a policy-over-options by extending the policy gradient theorem to options, but the approach was only tested on a toy problem.

In contrast, STRAW learns macro-actions and a policy over them in an end-to-end fashion from only the environment's reward signal and without resorting to explicit pseudo-rewards or hand-crafted subgoals.
The macro-actions are represented \emph{implicitly} inside the model, arising naturally from the interplay between action and commitment plans within the network. Our experiments demonstrate that the model scales to a variety of tasks from next character prediction in text to ATARI games.

\seckiny

\section{The model}
\seckiny
\label{sec:model}

STRAW is a deep recurrent neural network with two modules. The first module translates environment observations into an \emph{action-plan} -- a state variable which represents an explicit stochastic plan of future actions. STRAW generates \emph{macro-actions} by committing to the action-plan and following it without updating for a number of steps.
The second module maintains \emph{commitment-plan} -- a state variable that determines at which step the network terminates a macro-action and updates the action-plan. 
The action-plan is a matrix where one dimension corresponds to time and the other to the set of possible discrete actions. The elements of this matrix are proportional to the probability of taking the corresponding action at the corresponding time step. Similarly, the commitment plan represents the probabilities of terminating a macro-action at the particular step.
For updating both plans we use attentive writing technique~\cite{gregor2015draw}, which allows the network to focus on parts of a plan where the current observation is informative of desired outputs.
This section formally defines the model, we describe the way it is trained later in section~\ref{sec:learning}.

\begin{figure}[t]
\figskiny
\hspace{10mm} \includegraphics[scale=0.2]{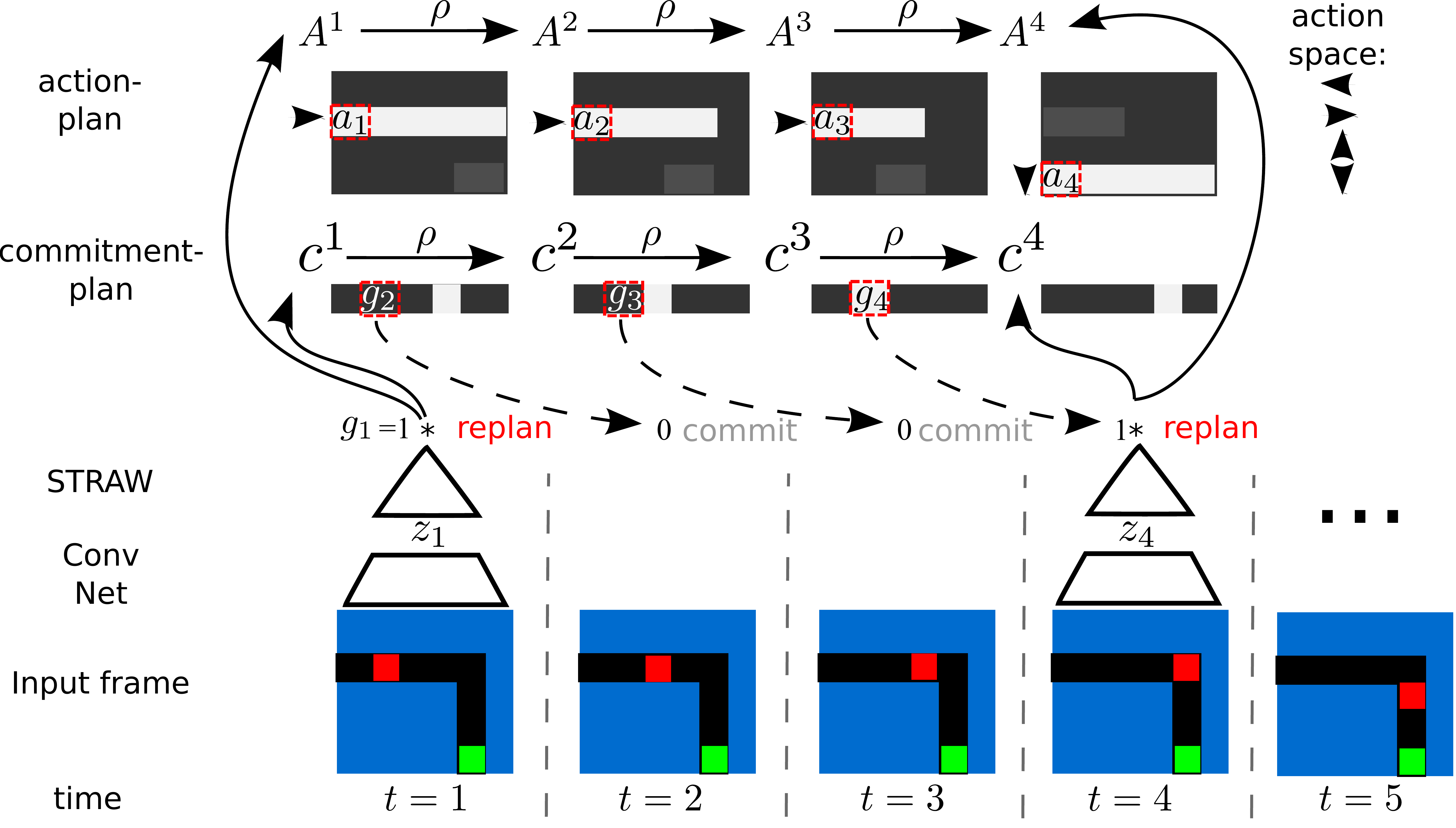}
  \caption{Schematic illustration of STRAW playing a maze navigation game. The input frames indicate maze geometry (black = corridor, blue = wall), red dot corresponds to the position of the agent and green to the goal, which it tries to reach. A frame is first passed through a convolutional network, acting as a feature extractor and then into STRAW. The top two rows depict the plans $\matr{A}$ and $\matr{c}$. Given a gate $g_t$, STRAW either updates the plans (steps 1 and 4) or commits to them. \label{fig:data_flow} \figskiny \figskiny }
\end{figure}

\parskiny
\paragraph{The state of the network} at time $t$ is comprised of matrices $\matr{A}^t\in R^{A \times T}$ and $\matr{c}^t\in R^{1 \times T}$. Matrix $\matr{A}^t$ is the \emph{action-plan}.
Each element $\matr{A}^t_{a,\tau}$ is proportional to the probability of outputting $a$ at time $t+\tau$. Here $A$ is a total number of possible actions and $T$ is a maximum time horizon of the plan.
To generate an action at time $t$, the first column of $\matr{A}^t$ (i.e. $\matr{A}^t_{\bullet 0}$) is transformed into a distribution over possible outputs by a $\mathit{SoftMax}$ function. This distribution is then sampled to generate the action $a_t$. 
Thereby the content of $\matr{A}^t$ corresponds to the plan of future actions as conceived at time $t$.
The single row matrix $\matr{c}^t$ represents the \emph{commitment-plan} of the network. 
Let $g_t$ be a binary random variable distributed as follows: $g_t \sim \matr{c}^{t-1}_1$. If $g_t=1$ then at this step the plans will be updated, otherwise $g_t=0$ means they will be committed to.
Macro-actions are defined as a sequence of outputs $\{a_t\}_{t_1}^{t_2-1}$ produced by the network between steps where $g_t$ is `on': i.e $g_{t_1}=g_{t_2}=1$ and $g_{t'}=0, \forall t_1<t'<t_2$. 
During commitment the plans are rolled over to the next step using the matrix \emph{time-shift} operator $\rho$, which shifts a matrix by removing the first column and appending a column filled with zeros to its rear. Applying $\rho$ to $\matr{A}^t$ or $\matr{c}^t$ reflects the advancement of time. Figure~\ref{fig:data_flow} illustrates the workflow. Notice that during commitment (step 2 and 3) the network doesn't compute the forward pass, thereby saving computation. 

\parskiny
\paragraph{Attentive planning.}
An important assumption that underpins the usage of macro-actions is that one observation reveals enough information to generate a sequence of actions. 
The complexity of the sequence and its length can vary dramatically, even within one environment. 
Therefore the network has to focus on the part of the plan where the current observation is informative of desired actions.
To achieve this, we apply differentiable attentive reading and writing operations~\cite{gregor2015draw}, where attention is defined over the temporal dimension. This technique was originally proposed for image generation, here instead it used to update the plans $\matr{A}^t$ and $\matr{c}^t$. In the image domain, the attention operates over the spatial extent of an image, reading and writing pixel values. Here it operates over the temporal extent of a plan, and is used to read and write action probabilities. The differentiability of the attention model~\cite{gregor2015draw} makes it possible to train with standard backpropagation.

An array of Gaussian filters is applied to the plan, yielding a ‘patch’ of smoothly varying location and zoom. Let $A$ be the total number of possible actions and $K$ be a parameter that determines the temporal resolution of the patch. A grid of $K\times A$ of one-dimensional Gaussian filters is positioned on the plan by specifying the co-ordinates of the grid center and the stride distance between adjacent filters.  The stride controls the ‘zoom’ of the patch; that is, the larger the stride, the larger an area of the original plan will be visible in the attention patch, but the lower the effective resolution of the patch will be. The filtering is performed along the temporal dimension only. Let $\psi$ be a vector of attention parameters, i.e.: grid position, stride, and standard deviation of Gaussian filters. We define the attention operations as follows:

\eqskiny
\begin{equation}
\matr{D}=\textit{write}(p, \psi^A_t); \ \ \
\beta_t = \textit{read}(\matr{A}^t, \psi^A_t)
\label{eq:read_write}
\end{equation}
\eqskiny

The \emph{write} operation takes in a patch $p \in R^{A \times K}$ and attention parameters $\psi$. It produces a matrix $\matr{D}$ of the same size as $\matr{A}^t$, which contains the patch $p$ scaled and positioned according to $\psi$.
Analogously the \emph{read} operation takes the full plan $\matr{A}^t$ together with attention parameters $\psi^A_t$ and outputs a read patch $\beta \in R^{A \times K}$, which is extracted from $\matr{A}^t$ according to $\psi^A_t$. 
We direct readers to~\cite{gregor2015draw} for details.

\parskiny
\paragraph{Action-plan update.} 

\begin{figure}[t]
\figskiny
\hspace{5mm} \includegraphics[scale=0.25]{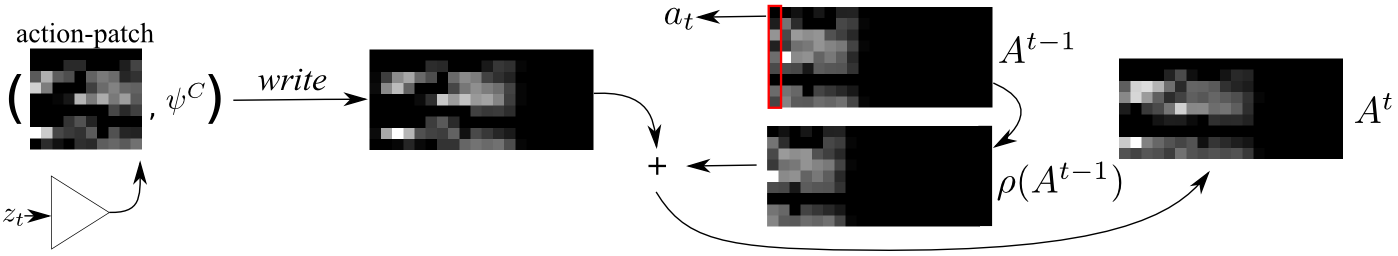}
  \caption{Schematic illustration of an action-plan update. Given $z_t$, the network produces an action-patch and attention parameters $\psi_t^A$. The \emph{write} operation creates an update to the action-plan by scaling and shifting the action-patch according to $\psi_t^A$. The update is then added to $\rho(\matr{A}^t)$. \figskiny \label{fig:attentive_write}}
\end{figure}

Let $z_t$ be a feature representation (e.g. the output of a deep convolutional network) of an observation $x_t$. 
Given $z_t$, $g_t$ and the previous state $\matr{A}^{t-1}$ STRAW computes an update to the action-plan using the Algorithm~\ref{alg:act_plan}. Here $f^\psi$ and $f^A$ are linear functions and $h$ is two-layer perceptron. Figure~\ref{fig:attentive_write} gives an illustration of an update to $\matr{A}^t$.

\begin{algorithm}[h]
\caption{Action-plan update}
\begin{algorithmic}
\State Input: $z_t, g_t, \matr{A}^{t-1}$
\State Output: $\matr{A}^t,a_t$
\If {$g_t = 1$}
\State Compute attention parameters $\psi^A_{t} = f^\psi(z_{t})$
\State Attentively read the current state of the action-plan $\beta_{t} = \textit{read}(\matr{A}^{t-1}, \psi^A_t);$
\State Compute intermediate representation $\xi_{t} = h([\beta_t,z_t])$
\State Update $\matr{A}^{t} = \rho(\matr{A}^{t-1})+g_t \cdot \textit{write}(f^A(\xi_t), \psi^A_t)$
\Else {\ // $g_t=0$}
\State Update $\matr{A}^{t} = \rho(\matr{A}^{t-1})$ \emph{//just advance time}
\EndIf
\State Sample an action $a_t \sim \textit{SoftMax}(\matr{A}^t_0)$
\end{algorithmic}
\label{alg:act_plan}
\end{algorithm}

\parskiny
\paragraph{Commitment-plan update.} Now we introduce a module that partitions the action-plan into macro-actions by defining the temporal extent to which the current action-plan $\matr{A}^t$ can be followed without re-planning. 
The commitment-plan $\matr{c}^t$ is updated at the same time as the action-plan, i.e. when $g_t=1$ or otherwise it is rolled over to the next time step by $\rho$ operator.
Unlike the planning module, where $\matr{A}^t$ is updated additively, $\matr{c}^t$ is overwritten completely using the following equations:

\eqskiny
\[ g_t \sim \matr{c}^{t-1}_1 \ \ \mbox{if } g_t=0 \mbox{ then } \matr{c}^t=\rho(\matr{c}^{t-1}) \mbox{ else }
 \psi^c_t = f^c([\psi^A_t,\xi_t])\]
\begin{equation}
 \matr{c}_{t} = \textit{Sigmoid}(\matr{b} + \textit{write}(e,\psi^c));
\label{eq:commit_replan}
\end{equation}
\eqskiny

Here the same attentive writing operation is used, but with only \emph{one} Gaussian filter for attention over $\matr{c}$. The patch $e$ is therefore just a scalar, which we fix to a high value ($40$ in our experiments). 
This high value of $e$ is chosen so that the attention parameters $\psi^c$ define a time step when re-planning is \emph{guaranteed} to happen. 
Vector $\matr{b}$ is of the same size as $d_t$ filled with a shared, learnable bias $b$, which defines the probability of re-planning earlier than the step implied by $\psi^c$.

Notice that $g_t$ is used as a multiplicative gate in algorithm~\ref{alg:act_plan}. This allows for retroactive credit assignment during training, as gradients from \emph{write} operation at time $t+\tau$ directly flow into the commitment module through state $\matr{c}^t$. Moreover, when $g_t=0$ only the computationally cheap operator $\rho$ is invoked. Thereby more commitment significantly saves computation.

\seckiny

\subsection{Structured exploration with macro-actions}
\label{sec:structured_exp}

The architecture defined above is capable of generating macro-actions. This section describes how to use macro-actions for structured exploration. We introduce STRAW-explorer (STRAWe), a version of STRAW with a noisy communication channel between the feature extractor (e.g. a convolutional neural network) and STRAW planning modules. 
Let $\zeta_t$ be the activations of the last layer of the feature extractor. We use $\zeta_t$ to regress the parameters of a Gaussian distribution $Q(z_t |\zeta_t) = N(\mu(\zeta_t),I\cdot\sigma(\zeta_t))$ from which $z_t$ is sampled. Here $\mu$ is a vector and $\sigma$ is a scalar. 
Injecting noise at this level of the network generates randomness on the level of plan updates that cover multiple time steps. This effect is reinforced by commitment, which forces STRAWe to execute the plan and experience the outcome instead of rolling the update back on the next step. In section~\ref{sec:experiments} we demonstrate how this significantly improves score on games like Frostbite and Ms. Pacman.

\seckiny

\section{Learning}
\seckiny
\label{sec:learning}
The training loss of the model is defined as follows:

\eqskiny
\begin{equation}
\mathcal{L}= \sum_{t}^T \left({ L^\textit{out}(\matr{A}^t) + \mathbf{1}_{g_t} \cdot  \alpha \textit{KL}(Q(z_t|\phi(x_t))|P(z_t)) + 
\lambda \matr{c}_t[t] }\right),
\end{equation}
\eqskiny

where $L^\textit{out}$ is a domain specific differentiable loss function defined over the network's output. For supervised problems, like next character prediction in text, $L^\textit{out}$ can be defined as negative log likelihood of the correct output. We discuss the reinforcement learning case later in this section.
The two extra terms are regularisers. The first is a cost of communication through the noisy channel, which is defined as KL divergence between latent distributions $Q(z_t|\phi(x_t))$ and some prior $P(z_t)$. Since the latent distribution is a Gaussian (sec.~\ref{sec:structured_exp}), a natural choice for the prior is a Gaussian with a zero mean and standard deviation of one.
The last term penalizes re-planning and encourages commitment.

For reinforcement learning we consider the standard setting where an agent is interacting with an environment in discrete time. At each step $t$, the agent observes the state of the environment $x_t$ and selects an action $a_t$ from a finite set of possible actions. The environment responds with a new state $x_{t+1}$ and a scalar reward $r_t$. The process continues until the terminal state is reached, after which it restarts. The goal of the agent is to maximize the discounted return 
$R_t= \sum_{k=0}^{\infty} \alpha^k r_{t+k+1}$. The agent's behaviour is defined by its policy $\pi$ -- mapping from state space into action space.
STRAW produces a distribution over possible actions (a stochastic policy) by passing the first column of the action-plan $\matr{A}^t$ through a SoftMax function: $\pi(a_t|x_t;\theta) = \textit{SoftMax}(\matr{A}^t_{\bullet 0})$. An action is then produced by sampling the output distribution.

We use a recently proposed Asynchronous Advantage Actor-Critic (A3C) method~\cite{mnih2016asynchronous}, which directly optimizes the policy of an agent. A3C requires a value function estimator $V(x_t)$ for variance reduction. 
This estimator can be produced in a number of ways, for example by a separate neural network. 
The most natural solution for our architecture would be to create value-plan containing the estimates. 
To keep the architecture simple and efficient, we simply add an auxiliary row to the action plan which corresponds to the value function estimation.
It participates in attentive reading and writing during the update, thereby sharing the temporal attention with action-plan.
The plan is then split into action part and the estimator before the $\mathit{SoftMax}$ is applied and an action is sampled.
The policy gradient update for $L^{\textit{out}}$ in $\mathcal{L}$ is defined as follows:

\eqskiny
\begin{equation}
\nabla L^{\textit{out}} = \nabla_{\theta}\log\pi(a_t|x_t;\theta)( R_t - V(x_t;\theta)) + \beta \nabla_\theta H(\pi(a_t|x_t;\theta))
\label{eq:learning}
\end{equation}
\eqskiny

Here $H(\pi(a_t|x_t;\theta))$ is entropy of the policy, which stimulates the exploration of primitive actions.
The network contains two random variables -- $z_t$ and $g_t$, which we have to pass gradients through. For $z_t$ we employ the re-parametrization trick~\cite{kingma2014auto,rezende2014stochastic}. For $g_t$, we set $\nabla\matr{c}^{t-1}_1 \equiv \nabla g_t$ as proposed in~\cite{bengio2013estimating}.
\seckiny

\section{Experiments}
\label{sec:experiments}
\seckiny

The goal of our experiments was to demonstrate that STRAW learns meaningful and useful macro-actions. 
We use three domains of increasing complexity: supervised next character prediction in text~\cite{graves2013generating}, 2D maze navigation and ATARI games~\cite{bellemare-ale}.

\subseckiny
\subsection{Experimental setup}
\subseckiny
\paragraph{Architecture.} 
The read and write patches are $A \times 10$ dimensional, and $h$ is a 2 layer perceptron with 64 hidden units. The time horizon $T=500$. For STRAWe (sec.~\ref{sec:structured_exp}) the Gaussian distribution for structured exploration is 128-dimensional. Ablative analysis for some of these choices is provided in section~\ref{subsec:ablative}. Feature representation of the state space is particular for each domain. For 2D mazes and ATARI it is a convolutional neural net (CNN) and it is an LSTM~\cite{hochreiter1997lstm} for text. We provide more details in the corresponding sections.

\parskiny \parskiny
\paragraph{Baselines.} The experiments employ two baselines: a simple feed forward network (FF) and a recurrent LSTM network. FF directly regresses the action probabilities and value function estimate from feature representation. The LSTM~\cite{hochreiter1997lstm} architecture is a widely used recurrent network and it was demonstrated to perform well on a suite of reinforcement learning problems~\cite{mnih2016asynchronous}. It has 128 hidden units and its inputs are the feature representation of an observation and the previous action of the agent. Action probabilities and the value function estimate are regressed from its hidden state.

\parskiny \parskiny
\paragraph{Optimization.} We use the A3C method~\cite{mnih2016asynchronous} for all reinforcement learning experiments. It was shown to achieve state-of-the-art results on several challenging benchmarks~\cite{mnih2016asynchronous}.
We cut the trajectory and run backpropagation through time~\cite{mozer1989focused} after 40 forward passes of a network or if a terminal signal is received. 
The optimization process runs 32 asynchronous threads using shared RMSProp. There are 4 hyper-parameters in STRAW and 2 in the LSTM and FF baselines. For each method, we ran 200 experiments, each using randomly sampled hyperparameters. Learning rate and entropy penalty were sampled from a $\mathit{LogUniform}(10^{-4},10^{-3})$ interval. Learning rate is linearly annealed from a sampled value to 0. To explore STRAW behaviour, we sample coding cost $\alpha \sim \mathit{LogUniform}(10^{-7},10^{-4})$ and replanning penalty $\lambda \sim \mathit{LogUniform}(10^{-6},10^{-2})$.  For stability, we clip the advantage $R_t-V(x_t;\theta)$ (eq.~\ref{eq:learning}) to $[-1,1]$ for all methods and for STRAW(e) we do not propagate gradients from commitment module into planning module through $\psi^A_T$ and $\xi_t$. We define a training epoch as one million observations.

\begin{figure}[h]
\figskiny
\hspace{-7mm} \includegraphics[scale=0.3]{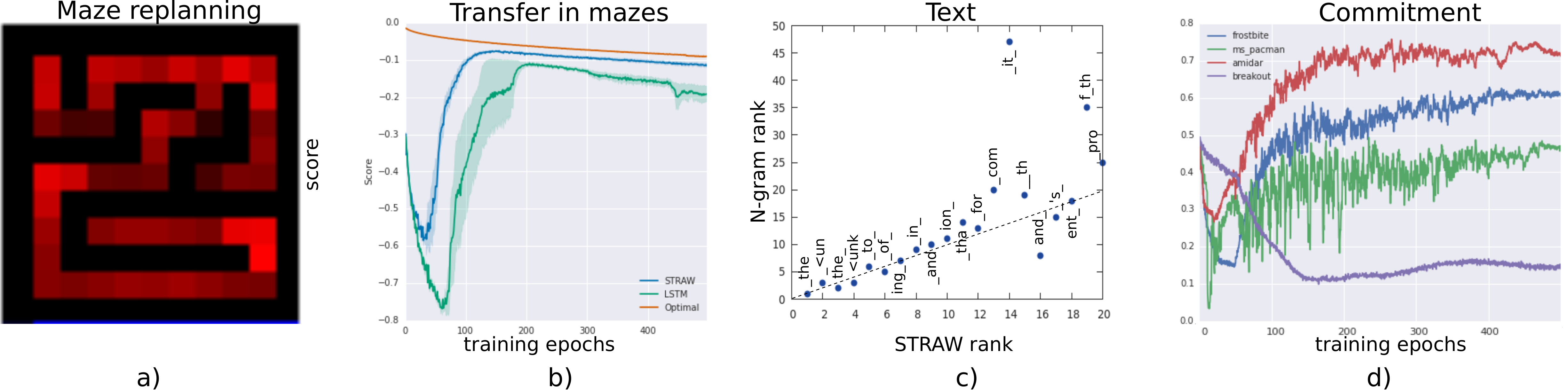}
\figskiny
  \caption{Detailed analysis. a) re-planning in random maze b) transfer to farther goals c) learned macro-actions for next character prediction d) commitment on ATARI. \figskiny \label{fig:extras} }
\end{figure}

\subseckiny
\subsection{Text}
\subseckiny
STRAW is a general sequence prediction architecture. To demonstrate that it is capable of learning output patterns with complex structure we present a qualitative experiment on next character prediction using Penn Treebank dataset~\cite{marcus1993ptb}. Actions in this case correspond to emitting characters and macro-actions to their sequences.
For this experiment we use an LSTM, which receives a one-hot-encoding of 50 characters as input. A STRAW module is connected on top. We omit the noisy Gaussian channel, as this task is fully supervised and does not require exploration. The actions now correspond to emitting characters. The network is trained with stochastic gradient descent using supervised negative log-likelihood loss (sec.~\ref{sec:learning}).
At each step we feed a character to the model which updates the LSTM representation, but only update the STRAW plans according to the commitment plan $\matr{c}^t$. For a trained model we record macro-actions -- the sequences of characters produced when STRAW is committed to the plan.
If STRAW adequately learns the structure of the data, then its macro-actions should correspond to common n-grams. Figure~\ref{fig:extras}.c plots the 20 most frequent macro-action of length 4 produced by STRAW. On x-axis is the rank of the frequency at which macro-action is used by STRAW, on y-axis is it's actual frequency rank as an 4-gram. Notice how STRAW correctly learned to predict frequent 4-grams such as 'the', 'ing', 'and' and so on.

\subseckiny
\subsection{2D mazes}
\subseckiny
To investigate what macro-actions our model learns and whether they are useful for reinforcement learning we conduct an experiment on a random 2D mazes domain. 
A maze is a grid-world with two types of cells -- walls and corridors. One of corridor cells is marked as a goal. The agent receives a small negative reward of $-r$ every step, and double that if it tries to move through the wall. It receives as small positive reward $r$ when it steps on the goal and the episode terminates. Otherwise episode terminates after 100 steps. Therefore to maximize return an agent has to reach the goal as fast as possible. In every training episode the position of walls, goal and the starting position of the agent are randomized. An agent fully observes the state of the maze. The remainder of this section presents two experiments in this domain. 
Here we use STRAWe version with structured exploration. For feature representation we use a 2-layer CNN with 3x3 filters and stride of 1, each followed by a rectifier nonlinearity.

In our first experiment we train a STRAWe agent on an 11 x 11 random maze environment. 
We then evaluate a trained agent on a novel maze with fixed geometry and only randomly varying start and goal locations. The aim is to visualize the positions in which STRAWe terminates macro-actions and re-plans. Figure~\ref{fig:extras}.a shows the maze, where red intensity corresponds to the ratio of re-planning events at the cell normalized with the total amount of visits by an agent. Notice how corners and areas close to junctions are highlighted. This demonstrates that STRAW learns an adequate temporal abstractions in this domain. In the next experiment we test whether these temporal abstractions are useful. 

The second experiment uses a larger 15 x 15 random mazes. If the goal is placed arbitrarily far from an agent's starting position, then learning becomes extremely hard and neither LSTM nor STRAWe can reliably learn a good policy. We introduce a curriculum where the goal is first positioned very close to the starting location and is moved further away during the progress of training. More precisely, we position the goal using a random walk starting from the same point as an agent. We increase the random walks length by one every 2 million training steps, starting from 2. 
Although the task gets progressively harder, the temporal abstractions (e.g. follow the corridor, turn the corner) remain the same. If learnt early on, they should make adaptation easy.
The Figure~\ref{fig:extras}.b plots episode reward against training steps for STRAW, LSTM and the optimal policy given by Dijkstra algorithm. Notice how both learn a good policy after approximately 200 epochs, when the task is still simple. As the goal moves away LSTM has a strong decline in reward relative to the optimal policy. In contrast, STRAWe effectively uses macro-actions learned early on and stays close to the optimal policy at harder stages. This demonstrates that temporal abstractions learnt by STRAW are useful.

\begin{table}[h]
\vspace{-2mm}
   \caption{Comparison of STRAW, STRAWe, LSTM and FF baselines on 8 ATARI games. The score is averaged over top 5 agents for each architecture after 500 epochs of training. }
  \label{tab:ATARI_score}
  \centering
\begin{tabular}{lllllllll}
    \toprule
     & Frostbite     & Ms. Pacman & Q-bert & Hero & Crazy cl. & Alien & Amidar & Breakout \\
    \midrule
    STRAWe & \bf{8108} & \bf{6902} & \bf{23892} & 36931 & \bf{153327} & \bf{3230} & 2022 & 386\\
    STRAW     & 4394 &6730 &20933 &36734 &143803 &2626 & \bf{2223} &344\\
    \midrule
    LSTM     & 1281 &4227 &23121 &36072 &128932& 2984 &1392 & \bf{644}\\
    FF     &946 &2014 &22947 &\bf{39929} &115944 &2300 &1292 &106\\
    \bottomrule
  \end{tabular}
\end{table}

\vspace{-3mm}
\subsection{ATARI}
\subseckiny
This section presents results on a subset of ATARI games. All compared methods used the same CNN architecture, input preprocessing, and an action repeat of 4. For feature representation we use a CNN with a convolutional layer with 16 filters of size 8 x 8 with stride 4, followed by a convolutional layer with with 32 filters of size 4 x 4 with stride 2, followed by a fully connected layer with 128 hidden units.  All three hidden layers were followed by a rectifier nonlinearity. This is the same architecture as in~\cite{mnih2016asynchronous,mnih-dqn-2015}, the only difference is that in pre-processing stage we keep colour channels. 

We chose games that require some degree of planning and exploration as opposed to purely reactive ones: Ms. Pacman, Frostbite, Alien, Amidar, Hero, Q-bert, Crazy Climber. We also added a reactive Breakout game to the set as a sanity check. Table.~\ref{tab:ATARI_score} shows the average performance of the top 5 agents for each architecture after 500 epochs of training. Due to action repeat, here an epoch corresponds to four million frames (across all threads). STRAW or STRAWe reach the highest score on 6 out of 8 games. They are especially strong on Frostbite, Ms. Pacman and Amidar. On Frostbite STRAWe achieves more than $6\times$ improvement over the LSTM score.
Notice how structured exploration (sec.~\ref{sec:structured_exp}) improves the performance on all games but one. 
STRAW and STRAWe do perform worse than an LSTM on breakout, although they still achieves a very good score (human players score below 100). 
This is likely due to breakout requiring fast reaction and action precision, rather than planning or exploration.
FF baseline scores worst on every game apart from Hero, where it is the best. Although the difference is not very large, this is still a surprising result, which might be due to FF having fewer parameters to learn.

\begin{figure}[h]
\figskiny
 \includegraphics[scale=0.48]{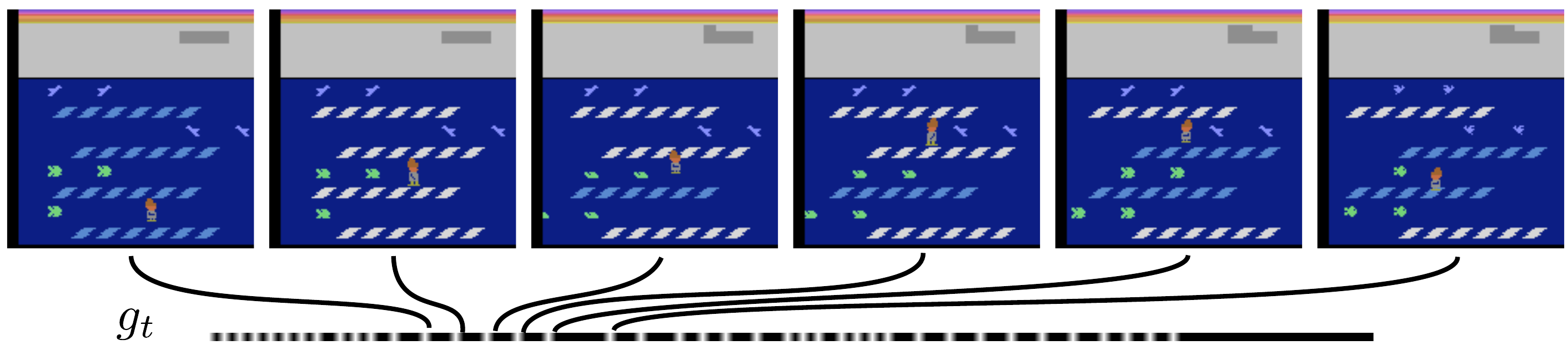}
 \figskiny
  \caption{Visualization of macro-actions on Frostbite. From top to bottom, the figure shows frames at which STRAW re-plans; $g_t$ -- white for 1, black for zero. Notice how macro-actions correspond to meaningful high-level actions like jumping from floe to floe and picking fish.  \label{fig:frostbite} \figskiny }
\end{figure}

\begin{figure}[h]

\hspace{-15mm}	\includegraphics[scale=0.3]{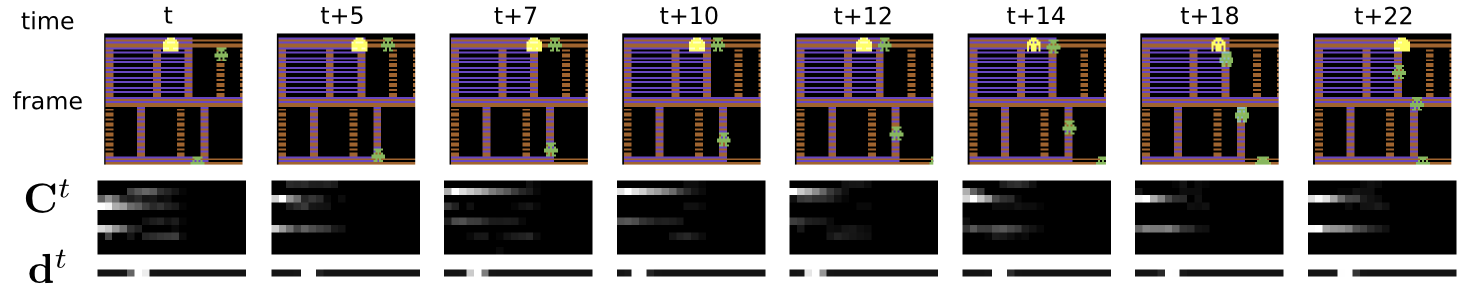}
\figskiny
  \caption{Re-planning behaviour in Amidar. The agent controls yellow figure, which scores points by exploring the maze, while avoiding green enemies. At time $t$ the agent has explored the area to the left and below and is planning to head right to score more points. At $t+7$ an enemy blocks the way and STRAW retreats to the left by drasticly changing the plan $\matr{A}^{t}$. It then resotres the original plan on step $t+14$ when the path is clear and heads to the right. Notice, that when the enemy is near ($t+7$ to $t+12$) it plans for smaller macro-actions --  $\matr{c}^t$ has a high value (white spot) closer to the origin. \label{fig:amidar} \figskiny \figskiny }
\end{figure}

Examples of macro-actions learned by STRAWe on Frostbite are shown in figure~\ref{fig:frostbite}. Notice that macro-actions correspond to jumping from floe to floe and picking up fish. Figure~\ref{fig:amidar} demonstrates re-planning behaviour on Amidar. In this game, agent explores a maze with a yellow avatar. It has to cover all of the maze territory without colliding with green figures (enemies). Notice how the STRAW agent changes its plan as an enemy comes near and blocks its way. It backs off and then resumes the initial plan when the enemy takes a turn and danger is avoided. Also, notice that when the enemy is near, agent plans for shorter macro-actions as indicated by commitment-plan $\matr{c}^t$. Figure~\ref{fig:extras}.d shows the percentage of time the best STRAWe agent is committed to a plan on 4 different games. As training progresses, STRAWe learns to commit more and converges to a stable regime after about 200 epochs. The only exception is breakout, where meticulous control is required and is beneficial to re-plan often. This shows that STRAWe is capable of adapting to the environment and learns temporal abstractions useful for each particular case.

\begin{figure}[h+]
\figskiny \figskiny
\begin{center}
    \subfigure{\label{subfig:AW}\includegraphics[scale=0.3]{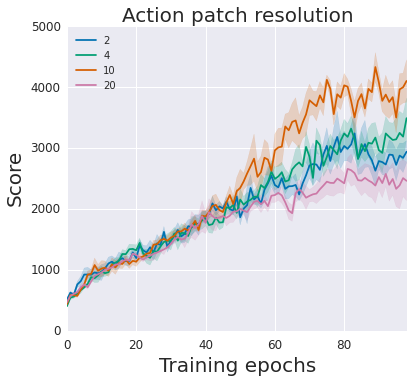}}
    \subfigure {\label{subfig:NH}\includegraphics[scale=0.3] {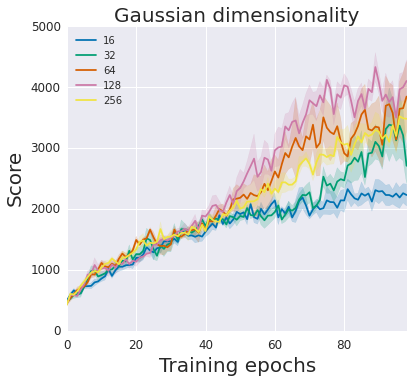} }
    \subfigure {\label{subfig:sleep}\includegraphics[scale=0.3] {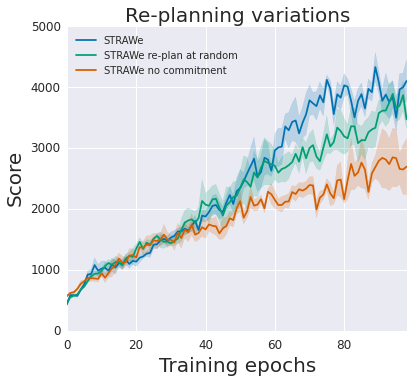} }
\end{center}
\figskiny
\caption{ Ablative analysis on Ms. Pacman game. Figures plot episode reward against seen frames for different configurations of STRAW. From left to right: varying action patch size, varying code length, varying replanning modules. Notice, that models here are trained for only 100 epochs, unlike models in Table~\ref{tab:ATARI_score} that were trained for 500 epochs. \figskiny \label{fig:ablative}}

\end{figure}

\subseckiny
\subsection{Ablative analysis}
\label{subsec:ablative}
\subseckiny
Here we examine different design choices that were made and investigate their impact on final performance. Figure~\ref{fig:ablative} presents the performance curves of different versions of STRAW on Ms. Pacman game trained for 100 epochs. From left to right, the first plot shows STRAW performance given different resolution of the action patch. The higher it is, the more complex is the update that STRAW can generate for the action plan. The second plot shows the influence of the Gaussian channel's dimensionality, which is used for structured exploration (sec.~\ref{sec:structured_exp}). The results show that higher dimensionality is generally beneficial, but these benefits rapidly saturate. In the third plot we investigate different possible choices for the re-planning mechanism: we compare STRAW with two simple modifications, one re-plans at every step, the other commits to the plan for a random amount of steps between 0 and 4.  Re-planning at every step is not only less elegant, but also much more computationally expensive and less data efficient. The results demonstrate that learning when to commit to the plan and when to re-plan is beneficial.

\seckiny
\section{Conclusion}
\label{sec:conclusion}
\seckiny
We have introduced the STRategic Attentive Writer (STRAW) architecture, and demonstrated its ability to implicitly learn useful temporally abstracted macro-actions in an end-to-end manner. Furthermore, STRAW advances the state-of-the-art on several challenging Atari domains that require temporally extended planning and exploration strategies, and also has the ability to learn temporal abstractions in general sequence prediction. As such it opens a fruitful new direction in tackling an important problem area for sequential decision making and AI more broadly. 

\seckiny
\section{Acknowledgements}
\label{sec:acknowledgements}
We thank David Silver, Joseph Modayil and Nicolas Heess for many helpful discussions, suggestions and comments on the paper.
\seckiny

\small

\setlength{\bibsep}{2pt plus 0.5ex}
\bibliography{deeprl}

\end{document}